# SlimYOLOv3: Narrower, Faster and Better for Real-Time UAV Applications

Pengyi Zhang, Yunxin Zhong, Xiaoqiong Li

School of Life Science

Beijing Institute of Technology

## Abstract

Drones or general Unmanned Aerial Vehicles (UAVs), endowed with computer vision function by on-board cameras and embedded systems, have become popular in a wide range of applications. However, real-time scene parsing through object detection running on a UAV platform is very challenging, due to limited memory and computing power of embedded devices. To deal with these challenges, in this paper we propose to learn efficient deep object detectors through channel pruning of convolutional layers. To this end, we enforce channel-level sparsity of convolutional layers by imposing L1 regularization on channel scaling factors and prune less informative feature channels to obtain "slim" object detectors. Based on such approach, we present SlimYOLOv3 with fewer trainable parameters and floating point operations (FLOPs) in comparison of original YOLOv3 (Joseph Redmon et al., 2018) as a promising solution for real-time object detection on UAVs. We evaluate SlimYOLOv3 on VisDrone2018-Det benchmark dataset; compelling results are achieved by SlimYOLOv3 in comparison of unpruned counterpart, including ~90.8% decrease of FLOPs, ~92.0% decline of parameter size, running ~2 times faster and comparable detection accuracy as YOLOv3. Experimental results with different pruning ratios consistently verify that proposed SlimYOLOv3 with narrower structure are more efficient, faster and better than YOLOv3, and thus are more suitable for real-time object detection on UAVs. Our codes are made publicly available at https://github.com/PengyiZhang/SlimYOLOv3.

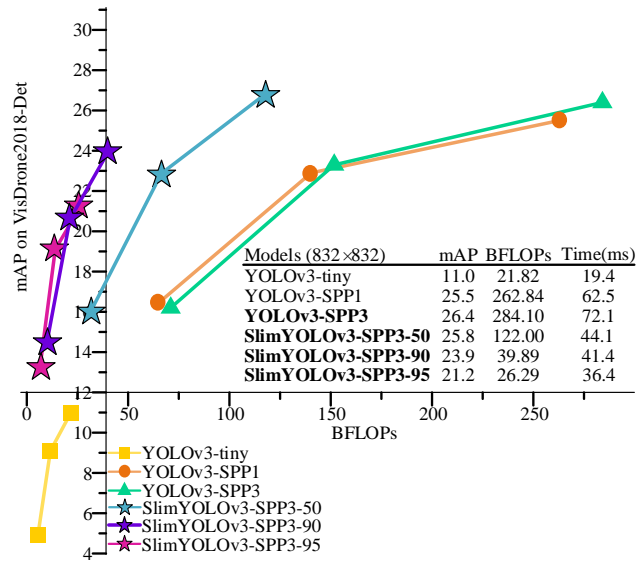

| Models (832×832) | mAP | BFLOPs | Time(ms) |
|---|---|---|---|
| YOLOv3-tiny | 11.0 | 21.82 | 19.4 |
| YOLOv3-SPP1 | 25.5 | 262.84 | 62.5 |
| **YOLOv3-SPP3** | **26.4** | **284.10** | **72.1** |
| **SlimYOLOv3-SPP3-50** | **25.8** | **122.00** | **44.1** |
| **SlimYOLOv3-SPP3-90** | **23.9** | **39.89** | **41.4** |
| **SlimYOLOv3-SPP3-95** | **21.2** | **26.29** | **36.4** |

Figure 1. Billion floating point operations (BFLOPs) versus accuracy (mAP) on *VisDrone2018-Det* benchmark dataset. Enabled by channel pruning, our SlimYOLOv3-SPP3 achieves comparable detection accuracy as YOLOv3 but only requires the equivalent floating point operations as YOLOv3-tiny. Such performance is very competitive in drone applications. Details are given in §5.

# 1. Introduction

Recently, drones or general Unmanned Aerial Vehicles (UAVs), endowed with computer vision function by on-board cameras and embedded systems, have been deployed in a wide range of applications, involving surveillance [1], aerial photography [2], and infrastructural inspection [3]. These applications require UAV platforms are able to sense environment, parse scene and react accordingly, of which the core part is scene parsing. Different drone applications require different levels of scene parsing, including recognizing what kinds of objects in the scene, locating where these objects are, and determining exact boundaries of each object. These scene parsing functions correspond to three basic research tasks in the field of computer vision, namely image classification, object detection and semantic (instance) segmentation. Visual object detection might be the most common one that is adopted as a basic functional module for scene parsing in UAV applications, and hence it has been the area of increasing interest. Due to the variety of open deployment environment, automatic scene parsing running on a UAV platform becomes highly demanding, which brings many new challenges to the object detection algorithms. These challenges mainly include: (1) how to deal with various variations (e.g., illumination, view, small sizes and ration) that object's visual appearance in aerial images commonly experiences; (2) how to deploy object detection algorithms on a UAV platform with limited memory and computing power; (3) how to balance the detection accuracy and real-time requirements. Object detection methods based on traditional machine learning and hand-crafted features are prone to failure when dealing with these variations. One competitive approach to addresses these challenges is object detectors based on deep learning techniques that are popularized in recent years.

Driven by the growth of computing power (e.g., Graphical Processing Units and dedicated deep learning chips) and the availability of large-scale labelled samples (e.g., ImageNet [4] and COCO [5]), deep neural network has been extensively studied due to its fast, scalable and end-to-end learning framework. Especially, compared with traditional shallow methods, Convolution Neural Network (CNN) [6] models have achieved significant improvements in image classification (e.g., ResNet[7] and DenseNet [8]), object detection (e.g., Faster R-CNN [9] and SSD [10]) and semantic segmentation (e.g., UNet [11] and Mask R-CNN [12]), etc. Since the beginning when CNN models were successfully introduced in object detection tasks (R-CNN, Ross Girshick et al., 2014) [13], this detection framework has attracted lots of research interest and many state-of-the-art object detectors based on CNN have been proposed in the past five years. Specifically, YOLO series models (Joseph Redmon et al. [14][15][16]) might be the most popular deep object detectors in practical applications as the detection accuracy and speed are well balanced. Despite that, the inference of these detectors still requires high-performance computing and large run-time memory footprint to maintain good detection performance; it brings high computation overhead and power consumption to on-board embedded devices of UAV platforms. Therefore, how to reduce floating point operations (FLOPs) and the size of trainable parameters without notably sacrificing detection precision becomes an urgent problem to be solved when deploying deep object detectors on UAVs. Model pruning methods is one promising approach to achieve this goal.

A typical deep learning pipeline briefly involves designing network structures, fine-tuning hyper-parameters, training and evaluating network. The majority of popular network structures (e.g., ResNet and DenseNet) are designed manually, in which the importance of each component cannot be determined before training. During the training process, network can learn the importance of each component through adjusting the weights in trainable layers automatically. Consequently, some connections and computations in the network become redundant or non-critical and hence can be removed without significant degradation in performance [17]. Based on this assumption, many model pruning methods have been designed recently to simplify deep models and facilitate the deployment of deep models in practical applications. Channel pruning is a coarse-grained but effective approach, and more importantly, it is convenient to implement the pruned models just by modifying the number of corresponding channel (or filter) in configuration files. A fine-tuning operation is subsequently performed on pruned models to compensate potentially temporary degradation. We empirically argue that deep object detectors designed by experts manually might exist inherent redundancy in feature channels, thus making it possible to reduce parameter size and FLOPs through channel pruning.

In this paper, we propose to learn efficient deep object detectors through performing channel pruning on convolutional layers. To this end, we enforce channel-level sparsity of convolutional layers by imposing L1 regularization on channel scaling factors and prune the less informative feature channels with small scaling factors to obtain "slim" object detectors. Based on such approach, we further present

SlimYOLOv3 with fewer trainable parameters and lower computation overhead in comparison of original YOLOv3 [16] as a promising solution for real-time object detection on UAVs. YOLOv3 is initially trained with channel-level sparsity regularization; sequentially, SlimYOLOv3 is obtained by pruning feature channels to a certain ratio according to their scaling factors in YOLOv3; SlimYOLOv3 is finally fine-tuned to compensate temporary degradation in detection accuracy. We evaluate SlimYOLOv3 on *VisDrone2018-Det* benchmark dataset [18]; SlimYOLOv3 achieves compelling results compared with its unpruned counterpart: ~90.8% decrease of FLOPs, ~92.0% decline of parameter size, running ~2 times faster and comparable detection accuracy as YOLOv3. Experimental results with different pruning ratios consistently verify that proposed SlimYOLOv3 with narrower structure are more efficient, faster and better than YOLOv3, and thus are more suitable for real-time object detection on UAVs.

## 2. Related Work

### 2.1 Deep Object Detector

Before R-CNN (Ross Girshick et al., 2014) [13] was proposed, object detection was used to be treated as a classification problem through sliding windows on the images. These traditional methods cannot deal with various variations of objects' appearance effectively. Combining selective search and CNN models, R-CNN achieved notable improvements in object detection tasks in comparison of shallow methods. Since then, deep object detectors have attracted lots of research interest; many state-of-the-art deep object detectors have been proposed in the past five years, including SPP-net[19], Fast R-CNN[20], Faster R-CNN [9], R-FCN [21], RetinaNet [22], SSD [10], YOLO [14], YOLOv2 (YOLO9000) [15] and YOLOv3 [16], etc. According to whether extra region proposal modules are required, these deep object detectors can be simply divided into two categories, i.e., two-stage and single-stage detectors.

**Two-stage detectors**. Two-stage detectors represented by R-CNN series models mainly consist of three parts: (1) backbone network, (2) region proposal module, and (3) detection header. First, region proposal modules generate large numbers of region proposals that likely contain objects of interest; sequentially, detection headers classify these proposals to retrieve their categories and perform position regression to locate objects precisely. Detection accuracy and real-time performance of two-stage object detectors have been increasingly optimized through several major improvements in region proposal methods (e.g., selective search [13] and region proposal networks [9], etc.), deep feature computing methods of region proposal (spatial pyramid pooling [19], ROI pooling [9], ROI align [12], etc.) and backbone networks (VGG, ResNet [7], feature pyramid network [23], etc.). Two-stage detectors resort to region proposals of high quality generated by region proposal modules to obtain a good detection accuracy. However, the inference of two-stage detectors with these region proposals requires huge computation and run-time memory footprint, thus making detection relatively slow.

**Single-stage detectors**. In contrast, single-stage detectors represented by YOLO series models, SSD and RetinaNet utilize predefined anchors that densely cover spatial positions, scales and aspect ratios across an image instead of using extra branch networks (e.g., region proposal network). In other words, single-stage detectors directly treat object detection as regression problems by taking input images and learning category probabilities and bounding box coordinates relative to predefined anchors. Encapsulating all computations in a single network, single-stage detectors are more likely to run faster than two-stage detectors. Amongst these single-stage detectors, YOLO series models might be the fastest object detection algorithms with state-of-the-art detection accuracy and hence become one of the most popular deep object detectors in practical applications. The real-time performance of YOLO series models reported in the literatures are evaluated on powerful Graphical Processing Units (GPU) cards with high-performance computing capacity. When deploying on a UAV platform with limited computing capacity, it will be very challenging to balance detection performance and high computation overhead. In this paper, we propose to learn an efficient YOLOv3 model, i.e., SlimYOLOv3, through channel pruning of convolutional layers to deal with this challenge.

### 2.2 Model pruning

When deploying a deep model on resource-limited devices, model compression is a useful tool for researchers to rescale the resource consumption required by deep models. Existing model compression methods mainly include model pruning [17][24], knowledge distillation [25], parameter quantization [26] and dynamic computation [27], etc. In this section, we specifically discuss model pruning methods.

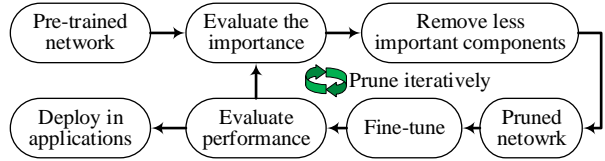

Figure 2. A representative procedure of incremental model pruning. There exists four iterative steps: (1) evaluating importance of each component in a pre-trained deep model; (2) removing the components that are less important to model inference; (3) fine-tuning pruned model to compensate potentially temporary degradation in performance; (4) evaluating the fine-tuned model to determine whether pruned model is suitable for deployment. An incremental pruning strategy is preferred to prevent over-pruning.

A representative procedure of incremental model pruning is shown in Fig.2. The components removed from deep models in model pruning methods can be individual neural connections [28] or network structures [17][24]. Weight pruning methods prune the less important connections with small weights. It is conceptually easy to understand, but it is hard to store the pruned model and speed up due to the generated irregular network architecture. Technically, weight pruning might not be suitable for practical applications unless special software library or dedicated hardware is designed to support the pruned model. Unlike weight pruning, structured pruning is more likely to produce regular and tractable network architectures. To obtain structured unimportance for structured pruning, researchers resort to sparsity training with structured sparsity regularization involving structured sparsity learning [29] and sparsity on channel-wise scaling factors [17][24]. Liu et al. [24] proposed a simple but effective channel pruning approach called network slimming. They directly adopted the scaling factors in batch normalization (BN) layers as channel-wise scaling factors and trained networks with L1 regularization on these scaling factors to obtain channel-wise sparsity. Channel pruning is a coarse-grained but effective approach, and more importantly, it is convenient to implement the pruned models without the requirements of dedicated hardware or software. They applied network slimming methods to prune CNN-based image classifiers and notably reduced both model size and computing operations. In this paper, we follow Liu's work and extend it to be a coarse-grained method of neural architecture search for efficient deep object detectors.

## 3. SlimYOLOv3

Experts design network architectures for object detectors manually. There is no guarantee that each component plays an important role in forward inference. We propose to learn efficient deep object detectors through performing channel pruning on convolutional layers. Specifically, we aim to search a more compact and effective channel configuration of convolutional layers to help reduce trainable parameters and FLOPs. To this end, we apply channel pruning in YOLOv3 to obtain SlimYOLOv3 by following the procedure shown in Fig. 3.

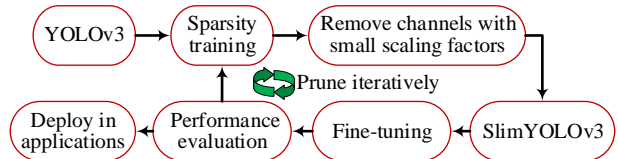

Figure 3. An iterative procedure of learning efficient deep object detector through sparsity training and channel pruning for SlimYOLOv3.

**YOLOv3 and YOLOv3-SPP3**. YOLOv3 makes an incremental improvement to the YOLO series models in object detection accuracy. First, YOLOv3 adopts a new backbone network, i.e., Darknet-53, as feature extractor. Darknet-53 uses more successive $3\times 3$ and $1\times 1$ convolutional layers than Darknet-19 in YOLOv2 and organizes them as residual blocks [7]. Hence, Darknet-53 is much more powerful than Darknet-19 but still more efficient than ResNet-101 [7]. Second, YOLOv3 predicts bounding boxes at three different scales by following the idea of feature pyramid network for object detection [23]. Three detection headers separately built on the top of three feature maps with different scales are responsible for detecting objects with different sizes. Each grid in the detection header is assigned with three different anchors, and thus predicts three detections that consist of 4 bounding box offsets, 1 objectiveness and $C$ class predictions. The final result tensor of detection header has a shape of $N \times N \times (3 \times (4 + 1 + C))$, where $N \times N$ denotes the spatial size of last convolutional feature map. In this paper, to enrich deep features with minimal modifications, we introduce spatial pyramid pooling (SPP) [19] module to YOLOv3. As shown in Fig. 4, the SPP module consists of 4 parallel maxpool layers with kernel sizes of $1\times 1$, $5\times 5$, $9\times 9$ and $13\times 13$. SPP module is able to extract multiscale deep features with different receptive fields and fuse them by concatenating them in the channel dimension of feature maps. The multiscale features obtained within same layer are expected to further improve detection accuracy of YOLOv3 with small computation cost. The additional feature channels introduced by SPP modules as well as extra FLOPs can be reduced and refined by channel pruning afterwards. In our experiments with *VisDrone2018-Det*, we integrate a SPP module in YOLOv3 between the 5[th] and 6[th] convolutional layers in front of each detection header to formulate YOLOv3-SPP3.

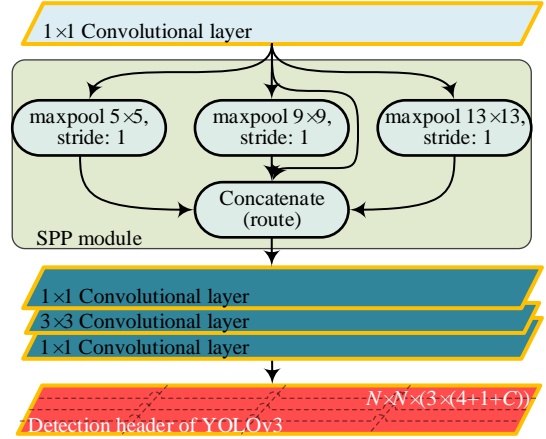

Figure 4. Architecture of SPP module used in YOLOv3-SPP3. We integrate a SPP module in YOLOv3 between the 5th and 6th convolutional layers in front of each detection header to formulate YOLOv3-SPP3.

**Sparsity training**. Channel-wise sparsity of deep models is helpful to channel pruning and describes the number of less important channels that are potential to be removed afterwards. To facilitate channel pruning, we assign a scaling factor for each channel, where the absolute values of scaling factors denote channel importance. Specifically, except for detection headers, a BN layer to accelerate convergence and improve generalization follows each convolutional layer in YOLOv3. BN layer normalize convolutional features using mini-batch statics, which is formulated as formula (1).

$$y = \gamma \times \frac{x - \bar{x}}{\sqrt{\sigma^2 + \varepsilon}} + \beta \quad (1)$$

where $\bar{x}$ and $\sigma^2$ are mean and variance of input features in a mini-batch, $\gamma$ and $\beta$ denotes trainable scale factor and bias. Naturally, we directly adopt the trainable scale factors in BN layers as indicators of channel importance. To discriminate important channels from unimportant channels effectively, we perform channel-wise sparsity training by imposing L1 regularization on $\gamma$. The training objective of sparsity training is given by formula (2).

$$L = loss_{yolo} + \alpha \sum_{\gamma \in \Gamma} f(\gamma) \quad (2)$$

Where $f(\gamma) = |\gamma|$ denotes L1-norm and $\alpha$ denotes penalty factor that balances the two loss terms. In our implementation, we use subgradient method to optimize the non-smooth L1 penalty term as Liu et al.[24] did.

**Channel pruning**. After sparsity training, we introduce a global threshold $\hat{\gamma}$ to determine whether a feature channel is to be pruned. The global threshold $\hat{\gamma}$ is set as *n*-th percentile of all $|\gamma|$ to control the pruning ratio. Besides, we also introduce a local safety threshold $\pi$ to prevent over-pruning on a convolutional layer and maintain the integrity of network connections. The local safety threshold $\pi$ is set in a layer-wise manner as *k*-th percentile of all $|\gamma|$ in a specific layer. We prune the feature channels whose scaling factors are smaller than the minimum of $\hat{\gamma}$ and $\pi$. In YOLOv3, several special connections between layers, e.g., the *route* layer and *shortcut* layer (Darknet [16]) are required to be treated carefully. During pruning process, we directly discard *maxpool* layer and *upsample* layer as they have nothing to do with channel number. Initially, we construct a pruning mask for all convolutional layers according to the global threshold $\hat{\gamma}$ and local safety threshold $\pi$. For a *route* layer, we concatenate pruning masks of its incoming layers in sequence and take the concatenated mask as its pruning mask. The *shortcut* layers in YOLOv3 play a similar role as residual learning in ResNet. Therefore, all the layers that have connections with *shortcut* layer are required to have a same channel number. To match the feature channels of each layer that are connected by *shortcut* layer, we iterate through the pruning masks of all connected layers and perform OR operation on these pruning masks to generate a final pruning mask for these connected layers.

**Fine-tuning**. After channel pruning, a fine-tuning operation is suggested to be performed on pruned models to compensate potentially temporary degradation. In fine-grained object detection tasks, detection performance is generally sensitive to channel pruning. Thus, fine-tuning is very important to make pruned model recover from potential degradation in performance. In our experiments with *VisDrone2018-Det*, we directly retrain SlimYOLOv3 using the same training hyper-parameters as the normal training of YOLOv3.

**Iteratively pruning**. As discussed in **section 2.2**, an incremental pruning strategy is preferred to prevent over-pruning. Over-pruning might lead to catastrophic degradation so that pruned model will never be recovered.

## 4. Experiments

We propose to learn efficient deep object detectors through pruning less important feature channels and further present SlimYOLOv3 with fewer trainable parameters and lower computation overhead for real-time object detection on UAVs. We empirically demonstrate the effectiveness of SlimYOLOv3 on *VisDrone2018-Det* benchmark dataset [18]. SlimYOLOv3 is implemented based on the publicly available Darknet [16] and a Pytorch implementation for YOLOv3 [30]. We use a Linux server with Intel(R) Xeon(R) E5-2683 v3 CPU @ 2.00GHz (56 CPUs), 64GB RAM, and four NVIDIA GTX1080ti GPU cards to train and evaluate models in our experiments.

### 4.1. Datasets
*VisDrone2018-Det* dataset consists of 7,019 static images captured by drone platforms in different places at different height [18]. The training and validation sets contain 6,471 and 548 images respectively. Images are labeled annotated with bounding boxes and ten predefined classes (i.e., pedestrian, person, car, van, bus, truck, motor, bicycle, awning-tricycle, and tricycle). All models in this paper are trained on training set and evaluated on validation set.

### 4.2. Models
**Baseline models**. We implement two YOLOv3 models, i.e., YOLOv3-tiny and YOLOv3-SPP1, as our baseline models. YOLOv3-tiny [16] is a tiny version of YOLOv3, and is much faster but less accurate. YOLOv3-SPP1 [16] is a revised YOLOv3, which has one SPP module in front of its first detection header. YOLOv3-SPP1 is better than original YOLOv3 on COCO dataset [5] in detection accuracy as reported in [16]. We thus take YOLOv3-SPP1 as a baseline of YOLOv3.

**YOLOv3-SPP3**. YOLOv3-SPP3 is implemented by incorporating three SPP modules in YOLOv3 between the 5th and 6th convolutional layers in front of three detection headers. YOLOv3-SPP3 is designed to further improve detection accuracy of baseline models.

**SlimYOLOv3**. We implement three SlimYOLOv3 models by setting the global threshold $\hat{\gamma}$ of channel pruning module as *50*-th percentile, *90*-th percentile and *95*-th percentile of all $|\gamma|$, corresponding to 50%, 90% and 95% pruning ratio respectively. The local safety threshold $\pi$ is empirically set as 90-th percentile of all $|\gamma|$ in each layer to keep at least 10% of channels unpruned in a single layer. We prune YOLOv3-SPP3 with these three pruning settings, and hence obtain SlimYOLOv3-SPP3-50, SlimYOLOv3-SPP3-90 and SlimYOLOv3-SPP3-95. Specifically, we iteratively prune YOLOv3-SPP3 2 times for SlimYOLOv3-SPP3-50 by following the iterative pruning procedure shown in Figure 3.

### 4.3. Training
**Normal training**. Following the default configurations in Darknet [16], we train YOLOv3-tiny, YOLOv3 and YOLOv3-SPP3 using SGD with the momentum of 0.9 and weight decay of 0.0005. We use an initial learning rate of 0.001 that is decayed by a factor of 10 at the iteration step of 70000 and 100000. We set the maximum training iteration as 120200 and use mini-batch size of 64. We set the size of input image as 416 for YOLOv3-tiny and 608 for YOLOv3 and YOLOv3-SPP3. Multiscale training is enabled by randomly rescaling the sizes of input images. We initialize the backbone networks of these three models with the weights pre-trained on ImageNet [4].

**Sparsity training**. We perform sparsity training for YOLOv3-SPP3 for 100 epochs. Three different values of penalty factor $\alpha$, i.e., 0.01, 0.001 and 0.0001, are used in our experiments. The remaining hyper-parameters of sparsity training are same as normal training.

**Fine-tuning**. We fine-tune SlimYOLOv3-SPP3-50, SlimYOLOv3-SPP3-90 and SlimYOLOv3-SPP3-

95 on training set. These models are initialized by the weights of pruned YOLOv3-SPP3. We use same hyper-parameters as in normal training to retrain SlimYOLOv3-SPP3-90 and SlimYOLOv3-SPP3-95 due to the possibility of aggressive pruning. For SlimYOLOv3-SPP3-50, we reduce maximum training iteration to 60200 and decay learning rate at the iteration step of 35000 and 50000 to fine-tune the pruned models.

It is to be noted that we use Darknet [16] to perform normal training and fine-tuning, while we use the Pytorch implementation [30] to perform sparsity training for convenience.

### 4.4. Evaluation metrics

We evaluate all these models based on the following 7 metrics: (1) precision, (2) recall, (3) mean of average precision (mAP) measured at 0.5 intersection over union (IOU), (4) F1-score, (5) model volume, (6) parameter size, (7) FLOPs and (8) inference time as frames per second (FPS). Specifically, the objectiveness confidence and non-maximum suppression threshold for all models in our experiments are set as 0.1 and 0.5 respectively. We run evaluation with no batch processing on one NVIDIA GTX1080ti GPU card using Darknet [16]. Besides, we evaluate all models with three different input sizes, including 416×416, 608×608 and 832×832.

## 5. Results and Discussions

We compare the detection performance of all models on validation set of VisDrone2018-Det dataset in Table 1 and Figure 1.

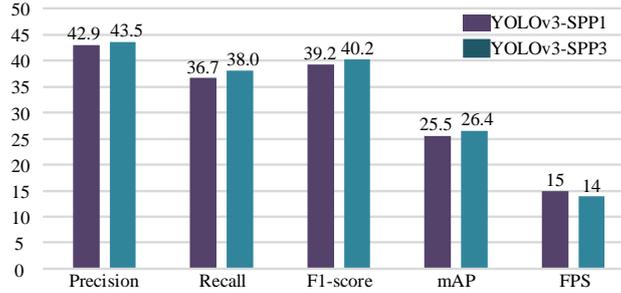

Figure 5: Performance comparison of YOLOv3-SPP1 and YOLOv3-SPP3 with input size of 832×832.

Table 1. Evaluation results of baseline models and pruned models.

| Model | Input size | Precision | Recall | F1-score | mAP | BFLOPS | FPS | Inference time (ms) | Parameters | Volume |
|---|---|---|---|---|---|---|---|---|---|---|
| YOLOv3-tiny | 416 | 19.5 | 10.5 | 13.1 | 4.9 | 5.46 | 134 | 7.5 | 8.7M | 33.1MB |
|  | 608 | 24.1 | 16.8 | 19.1 | 9.1 | 11.65 | 80 | 12.5 |  |  |
|  | 832 | 23.4 | 20.1 | 21.0 | 11.0 | 21.82 | 52 | 19.4 |  |  |
| YOLOv3-SPP1 | 416 | 39 | 24.5 | 29.5 | 16.5 | 65.71 | 46 | 21.7 | 62.6M | 239MB |
|  | 608 | 44.2 | 32.4 | 36.9 | 22.9 | 140.36 | 26 | 38.6 |  |  |
|  | 832 | 42.9 | 36.7 | 39.2 | 25.5 | 262.84 | 15 | 67.9 |  |  |
| YOLOv3-SPP3 | 416 | 36.6 | 24.9 | 29.1 | 16.2 | 71.03 | 40 | 24.8 | 63.9M | 243MB |
|  | 608 | 43.7 | 33.7 | 37.6 | 23.3 | 151.72 | 23 | 43.1 |  |  |
|  | 832 | 43.5 | 38 | 40.2 | 26.4 | 284.10 | 14 | 72.1 |  |  |
| SlimYOLOv3-SPP3-50 | 416 | 39.2 | 23.5 | 28.7 | 15.7 | 30.51 | 67 | 14.9 | 20.8M | 79.6MB |
|  | 608 | 45.6 | 32.1 | 37.1 | 22.6 | 65.17 | 39 | 25.6 |  |  |
|  | 832 | 45.9 | 36 | 39.8 | 25.8 | 122 | 23 | 44.1 |  |  |
| SlimYOLOv3-SPP3-90 | 416 | 32.2 | 21.6 | 24.4 | 14.5 | 9.97 | 67 | 14.8 | 8.0M | 30.6MB |
|  | 608 | 37.9 | 30.0 | 32.0 | 20.6 | 21.3 | 40 | 25.1 |  |  |
|  | **832** | **36.9** | **33.8** | **34.0** | **23.9** | **39.89** | **24** | **41.4** |  |  |
| SlimYOLOv3-SPP3-95 | 416 | 33.8 | 20.1 | 22.9 | 13.3 | 6.57 | 72 | 13.8 | 5.1M | 19.4MB |
|  | 608 | 37.3 | 28.2 | 30.1 | 19.1 | 14.04 | 41 | 24.1 |  |  |
|  | 832 | 36.1 | 31.6 | 32.2 | 21.2 | 26.29 | 28 | 36.4 |  |  |

**Effect of SPP modules**. With input sizes of 416×416 and 608×608, YOLOv3-SPP3 achieves comparable detection performance as YOLOv3-SPP1. With a larger input size, i.e., 832×832, YOLOv3-SPP3 outperforms YOLOv3-SPP1 by ~1% in mAP and F1-score as shown in Figure 5. It implies that SPP modules can help detectors extract useful multiscale deep features through different sizes of receptive fields in high-resolution input images. Correspondingly, the number of trainable parameters and FLOPs required by YOLOv3-SPP3 are slightly increased with the addition of SPP modules. The increased FLOPs (+21 BFLOPs) here are negligible in comparison of the decreased FLOPs (-244 BFLOPs with 90% pruning ratio) during channel pruning as shown in Figure 6.

**Effect of sparsity training**. During the sparsity training, we compute the histogram of scaling factors (absolute value) in all BN layers of YOLOv3-SPP3 to monitor change in the distribution of scaling factors. We visualize these histograms as well as the loss curves of training and validation sets in Figure 7. With the training progress, the number of smaller scaling factors increases while the number of larger factors decreases. Sparsity training is able to effectively reduce the scaling factors and thus make the feature channels of convolutional layers in YOLOv3-SPP3 sparse. However, sparsity training with a larger penalty factor, i.e., $\alpha = 0.01$, make the scaling factors decay so aggressive that models start failing with underfitting. In our experiments, we use the YOLOv3-SPP3 model trained with penalty factor $\alpha = 0.0001$ to perform channel pruning.

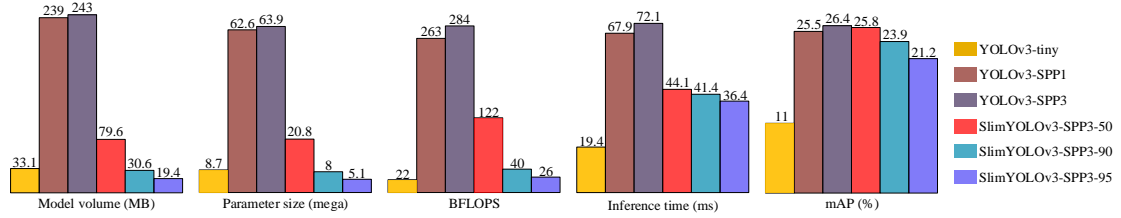

Figure 6. Comparison of baseline models and our SlimYOLOv3 models in model volume, parameter size, BLOPs, inference time and mAP score when input size is 832×832.

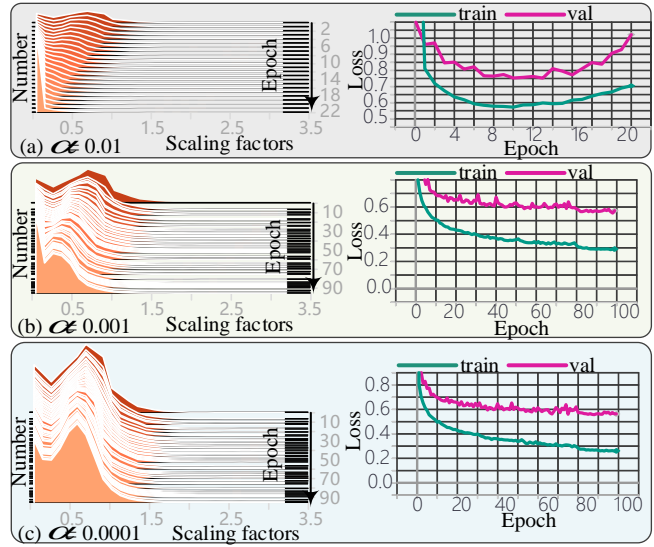

Figure 7. Histogram statistics of scaling factors in all BN layers (left) and loss curve of training and validation sets (right) during sparsity training of YOLOv3-SPP3 with three different values of $\alpha$ including 0.01, 0.001 and 0.0001. In (a), we terminate the sparsity training early when the model get stuck in underfitting.

**Effect of channel pruning**. In our experiments, we perform iterative pruning for SlimYOLOv3-SPP3-50 and aggressive pruning for SlimYOLOv3-SPP3-90 and SlimYOLOv3-SPP3-95 with three global thresholds corresponding to pruning ratio of 50%, 90% and 95% respectively. Compared with YOLOv3-SPP3, channel pruning with these three pruning ratio actually reduces FLOPs (when input size is 832×832) by 57.1%, 85.9% and 90.8%, decreases parameters size by 67.4%, 87.5% and 92.0%, and shrinks model volume by 67.5%, 87.4 and 92.0%. On the other hand, SlimYOLOv3-SPP3-90 and SlimYOLOv3-SPP3-95 are able to achieve comparable detection accuracy as YOLOv3 but requires even fewer trainable parameters than YOLOv3-tiny. Besides, the inference time (when input size is 832×832) evaluated on a NVIDIA GTX1080ti GPU card using Darknet [16] with no batch processing is reduced by 38.8%, 42.6% and 49.5% accordingly. It means SlimYOLOv3-SPP3 runs ~2 times faster than YOLOv3-SPP3. However, SlimYOLOv3-SPP3 runs much slower that YOLOv3-tiny with comparable FLOPs requirements as YOLOv3-tiny. One of the reasons for this phenomenon might be that YOLOv3-tiny has a shallower structure. During inference process, top layers in deep models always wait for the outputs from bottom layers to perform forward computation. Therefore, YOLOv3-tiny doesn't need to wait as longer as SlimYOLOv33-SPP3 to obtain the final detection outputs. We argue that this phenomenon implies that there might exist a bottleneck to improve real-time performance of deep object detectors through channel pruning.

**Analysis of detection accuracy**. As shown in Figure 1 and Table 1, the revised YOLOv3, i.e., YOLOv3-SPP3, achieves the best detection results but requires the most FLOPs at the meantime. In contrast, SlimYOLOv3-SPP3 models with even fewer trainable parameters than YOLOv3-tiny are able to obtain suboptimal detection results that are comparable with YOLOv3. Obviously, SlimYOLOv3-SPP3 is much better than YOLOv3-tiny in detection accuracy. Such results imply that with equivalent trainable parameters a deeper and narrower YOLOv3 model might be more powerful and effective than a shallower and wider YOLOv3 model. Besides, comparing SlimYOLOv3-SPP3-50 and SlimYOLOv3-SPP3-95 we can conclude that iterative pruning with a smaller pruning ratio are more prone to maintaining detection accuracy than aggressive pruning with a large pruning ratio. We produce visualized detection results of SlimYOLOv3-SPP3-95 and YOLOv3-SPP3 on a challenging frame captured by our drone as shown in Figure 8. Both of the two detectors are able to detect the majority of objects of interest precisely in this frame without significant difference.

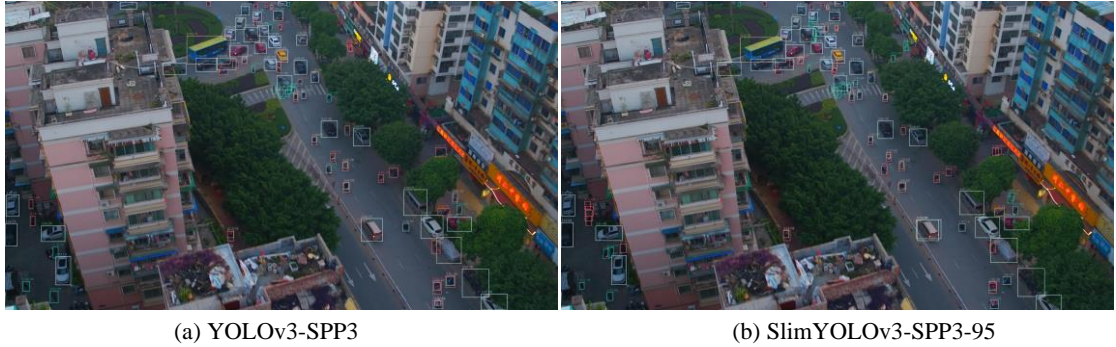

(a) YOLOv3-SPP3                      (b) SlimYOLOv3-SPP3-95

Figure 8: Visualized detection results of SlimYOLOv3-SPP3-95 and YOLOv3-SPP3 on a challenging frame captured by our drone.

**Limitations**. We have not made any modifications to both the training and inference of YOLOv3 expect for integrating SPP modules. However, *VisDrone2018-Det* is a very challenging dataset with high category imbalance. The category imbalance problem is not managed in purpose in our experiments. Category with a larger number of object instances dominates the optimization of detectors. Consequently, mAP score of this dominant category (i.e., *car*) is obviously higher than that of the categories (e.g., *bicycle*) with smaller number of instances as highlighted in Table 2 and Table 3. This issue occurs in both baseline models and pruned models, further leading to a significant decline in overall performance. Approaches for solving the category imbalance problem are left for future work to improve detection accuracy of both baseline models and pruned models.

Table 2. Detection performance of YOLOv3-SPP3 (832×832) for each category on validation set of *VisDrone2018-Det* dataset.

| Class | Images | Instances | Precision | Recall | F1-score | mAP |
|---|---|---|---|---|---|---|
| *pedestrian* | 548 | 8,840 | 46.6 | 38.0 | 46.8 | 33.2 |
| *people* | 548 | 5,120 | 41.8 | 35.7 | 38.5 | 20.3 |
| *bicycle* | 548 | 1,290 | 24.7 | 16.9 | 20.0 | **6.9** |
| *car* | 548 | 1,4100 | 68.8 | 78.2 | 73.2 | **70.1** |
| *van* | 548 | 1,980 | 43.7 | 39.4 | 41.4 | 27.4 |
| *truck* | 548 | 750 | 35.6 | 30.1 | 32.6 | 19.8 |
| *tricycle* | 548 | 1,040 | 35.5 | 25.7 | 29.9 | 12.8 |
| *awning-tricycle* | 548 | 532 | 23.4 | 14.5 | 17.9 | 6.6 |
| *bus* | 548 | 251 | 65.7 | 46.6 | 54.5 | 36.8 |
| *motor* | 548 | 4,890 | 49.0 | 46.1 | 47.5 | 30.4 |
| *overall* | 548 | 3,8800 | 43.5 | 38.0 | 40.2 | **26.4** |

Table 3. Detection performance of SlimYOLOv3-SPP3-95 (832×832) for each category on validation set of *VisDrone2018-Det* dataset.

| Class | Images | Instances | Precision | Recall | F1-score | mAP |
|---|---|---|---|---|---|---|
| *pedestrian* | 548 | 8,840 | 33.0 | 41.9 | 36.9 | 25.8 |
| *people* | 548 | 5,120 | 31.4 | 32.4 | 31.9 | 17.0 |
| *bicycle* | 548 | 1,290 | 14.4 | 10.3 | 12.0 | **2.7** |
| *car* | 548 | 1,4100 | 60.3 | 75.0 | 66.9 | **67.0** |
| *van* | 548 | 1,980 | 43.8 | 37.0 | 40.1 | 27.1 |
| *truck* | 548 | 750 | 26.8 | 27.6 | 27.2 | 16.4 |
| *tricycle* | 548 | 1,040 | 26.9 | 15.8 | 19.9 | 6.8 |
| *awning-tricycle* | 548 | 532 | 33.0 | 7.0 | 11.5 | 3.0 |
| *bus* | 548 | 251 | 55.9 | 28.3 | 37.6 | 22.8 |
| *motor* | 548 | 4,890 | 35.6 | 41.1 | 38.1 | 23.0 |
| *overall* | 548 | 3,8800 | 36.1 | 31.6 | 32.2 | **21.2** |

# 6. Conclusion

In this paper, we propose to learn efficient deep object detectors through channel pruning of convolutional layers. To this end, we enforce channel-level sparsity of convolutional layers by imposing L1 regularization on the channel scaling factors and prune the less informative feature channels with small scaling factors to obtain "slim" object detectors. Based on such approach, we further present SlimYOLOv3 with narrower structure and fewer trainable parameters than YOLOv3. Our SlimYOLOv3 is able to achieve comparable detection accuracy as YOLOv3 with significantly fewer FLOPs and run faster. As known to us all, power consumption is always positively correlated with FLOPs and low power consumption is generally required by drone applications to ensure endurance of drones. Therefore, we argue that SlimYOLOv3 is faster and better than original YOLOv3 for real-time UVA applications.

# Inference